\documentclass[letterpaper, 10pt, conference]{ieeeconf}      
\IEEEoverridecommandlockouts
\usepackage{graphicx}
\usepackage{subcaption}
\usepackage{amsmath,amssymb}
\usepackage{algorithmic}
\usepackage{algorithm}
\usepackage{xcolor}
\usepackage{tabularx}
\usepackage{booktabs}
\usepackage{multirow}
\usepackage{orcidlink}

\usepackage[nolist,nohyperlinks]{acronym}
\usepackage{hyperref}
\hypersetup{
    colorlinks=true,
    linkcolor=blue,
    citecolor=teal,
    filecolor=magenta,      
    urlcolor=cyan,
    pdftitle={main},
    pdfpagemode=FullScreen,
    }
\usepackage[backend=biber,style=ieee,sorting=none,abbreviate=true]{biblatex}

\begin{acronym}
    \acro{VAE}{Variational AutoEncoder}
    \acro{CVAE}{Conditional VAE}
    \acro{MSE}{Mean Square Error}
    \acro{MAE}{Mean Absolute Error}
    \acro{KLdiv}{Kullback–Leibler divergence}
    \acro{SMAPE}{Symmetric Mean Absolute Percentage Error}
    \acro{WMAPE}{Weighted Mean Absolute Percentage Error}
\end{acronym}

\title{\LARGE \bf Multi-modal perception for soft robotic interactions\\using generative models}

\author{Enrico Donato$^{1}$ \orcidlink{0000-0002-8844-5279}, Egidio Falotico$^{1}$ \orcidlink{0000-0001-8060-8080} and Thomas George Thuruthel$^{2}$ \orcidlink{0000-0003-0571-1672}
    \thanks{This work received funding from the European Union’s Horizon 2020 research and innovation program under grant agreement No. 863212 (PROBOSCIS project) and the Royal Society research grant RGS/R1/231472.}
    \thanks{$^{1}$E. Donato and E. Falotico are with The BioRobotics Institute, Sant'Anna School of Advanced Studies, 56025 Pontedera (PI), Italy and with the Departement of Excellence in Robotics \& AI, Sant'Anna School of Advanced Studies, 56125 Pisa, Italy {\tt\small \{e.donato, e.falotico\}@santannapisa.it}}%
    \thanks{$^{2}$T. George Thuruthel is with the Department of Computer Science, University College London, London, United Kingdom {\tt\small t.thuruthel@ucl.ac.uk}}%
}

\begin{document}
    \maketitle   

    \begin{abstract}
        Perception is essential for the active interaction of physical agents with the external environment. The integration of multiple sensory modalities, such as touch and vision, enhances this perceptual process, creating a more comprehensive and robust understanding of the world. Such fusion is particularly useful for highly deformable bodies such as soft robots. Developing a compact, yet comprehensive state representation from multi-sensory inputs can pave the way for the development of complex control strategies.

        This paper introduces a perception model that harmonizes data from diverse modalities to build a holistic state representation and assimilate essential information. The model relies on the causality between sensory input and robotic actions, employing a generative model to efficiently compress fused information and predict the next observation. We present, for the first time, a study on how touch can be predicted from vision and proprioception on soft robots, the importance of the cross-modal generation and why this is essential for soft robotic interactions in unstructured environments.  
    \end{abstract}

    \begin{keywords}
        Multi-modal Perception, Learning, Generative Models, Touch, Vision, Soft Robots
    \end{keywords}
    
    \section{Introduction}
        Being aware of oneself and one's surroundings requires information from various sources. Utilizing multiple senses enables a more comprehensive insight into the real world, especially when actions depend heavily on the perceptual input. This heightened state of awareness is imperative for the development of forthcoming robots that transcend mere reactivity to stimuli, actively engaging in the perception of the world through diverse sensory channels \cite{bajcsy2017activeperception}. The integration of multiple sensory modalities not only provides redundancy but also augments the resilience of the perceptual process \cite{tsai2019corruption}. Such enhancement is achievable through cross-modal inference \cite{li2019crossprediction}, wherein information from one modality is harnessed to derive inferences or draw conclusions across other modalities. This intricate process entails the amalgamation of cues from diverse sensory sources to form a more complete and coherent world representation.
        
        A perception model can be conceptualized as a cognitive framework that harmonizes data from diverse sensors and modalities, to build a compact, holistic, yet complete state representation of the experience \cite{tang2023fusiondeepl}. It is worth noting that despite the ongoing research into the construction of such a perception model, the means to do so remain an open question. Nevertheless, the potential impact of such a model on robot control has been well-established. During interactions, robots are reliant on the concurrent and complementary contributions of their various senses. These include vision for global inspection, touch for localized sensing, and proprioception for internal body representation \cite{nadon2018sensingmanipulation}. Several approaches have been proposed for the fusion of vision with touch \cite{lee2019touchvision, li2022seehearfeel, thuruthel2023visiontouch} and proprioception \cite{lee2020contachrichtasks, lee2021crossmodalcompensation}.

        \begin{figure}[t]
          \centering
          \includegraphics[width=\linewidth]{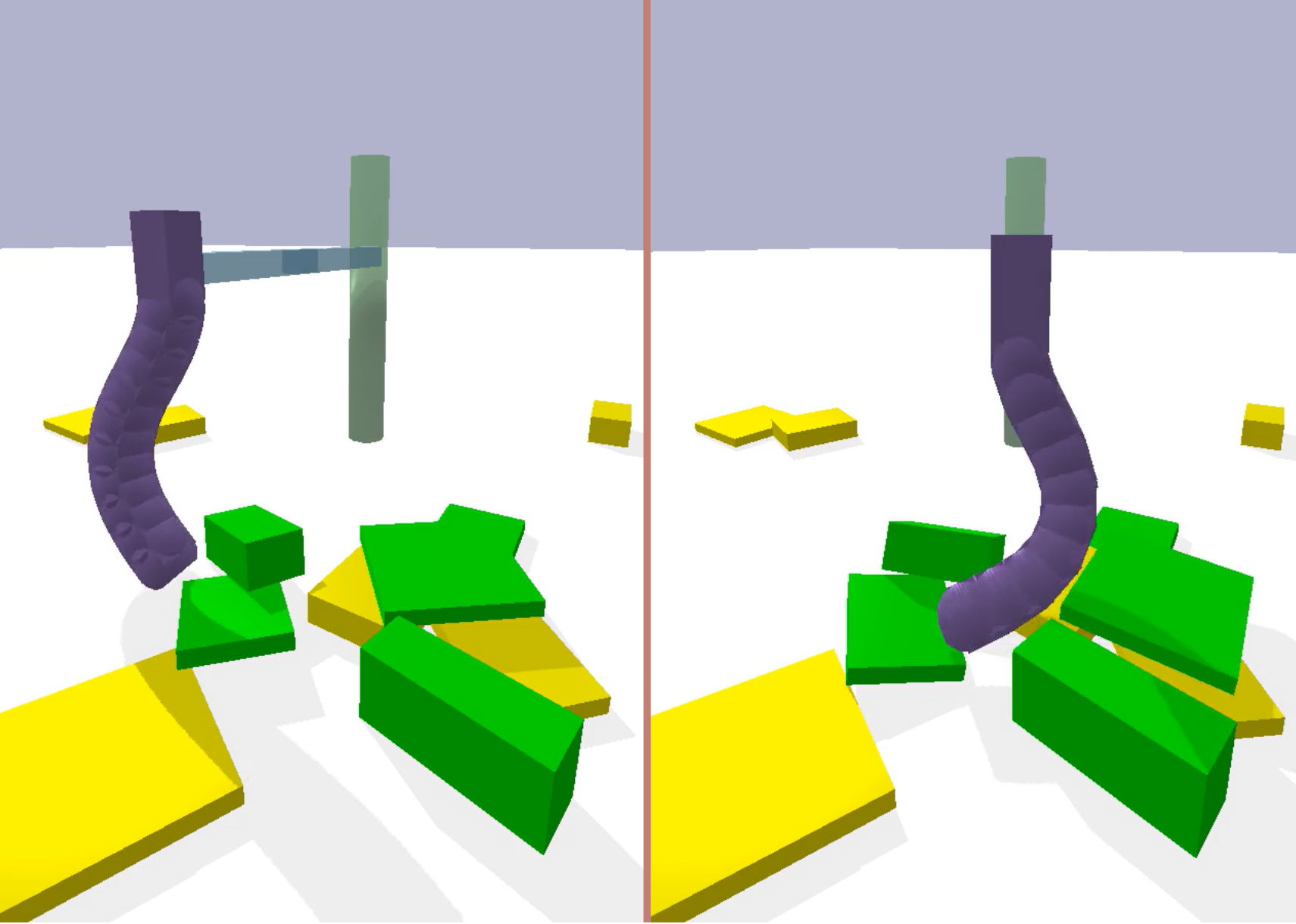}
          \caption{The simulated environment where a soft robot interacting in an unstructured environment attempts to combine visual-proprio feedback for compact state representation and tactile prediction.}
          \label{fig:task}
        \end{figure}
        
        Highly deformable bodies, such as soft robots, would benefit from a compact yet comprehensive state representation, especially when combining multi-sensory inputs or single-modality, distributed sensing along the body. Typically, modelling of soft or continuum robots often necessitates approximations, which limit the fidelity of the model when it comes to physical interactions \cite{armanini2023softrobotsmodeling}. Analytical models have been advanced to address their mathematical representation \cite{grazioso2019exactmodel,boyer2020strainmodel,alessi2023dynamicmodel}; however, the systems' complexity constrains their ease of applicability, particularly to incorporate multi-modal sensors for feedback control. Data-driven approaches \cite{kim2021mlsoftorbotics, laschi2023learningcontrol} can offer finite-dimensional models that entail lower computational costs. Additionally, model-free approaches have been proposed to discuss how to map sensing information to action straightly \cite{donato2022plantmovements,donato2023plantreaching}, but in contact-free tasks. Nevertheless, building useful state representation from multi-modal sensory inputs is an open challenge in the field of soft robotics for both analytical and learning-based methods when soft robots engage with their surroundings \cite{coevoet2017modelcontacthandling}. A compact multi-modal state representation, which assimilates relevant information, is vital for control in such environments.
        
        In this paper, we propose a learning architecture aimed at building a compact and informative state representation, by employing a predictive model conditioned on the forthcoming actions of the robot. By striving to extract the minimal amount of information necessary for predicting the evolution of the robot's body and its interaction with the environment, we aim to create a more efficient and streamlined perception model. These compact representations can then be employed as state representations for control tasks, significantly reducing the complexity of the control policy — a concept previously explored in the context of rigid systems \cite{lee2019touchvision}. The influence on soft robots will be even more profound, making them able to manage not only multi-modal sensing but also perceive localized information.
        
        Our implementation aims to forecast observations over time by leveraging the causal relationship between sensory input and robotic actions. Our fundamental hypothesis suggests that relying on a minimal state representation from multiple sensing modalities will enhance the predictive capabilities of the network. To achieve this, we employ a generative model to reduce sensing dimensionality and enable conditioned reconstruction, thereby achieving the most efficient compression of the fused information. The paper will additionally contribute to the analysis of cross-modal sensing generation, offering insights into touch prediction in relation to vision sensing.

        We first describe how the predictive model can be realized through the use of a generative model and delineate the benchmarking process for the fusion and compression of the state representation using suitable architectures. Subsequently, we introduce a simulation environment for data collection, as depicted in Figure \ref{fig:task}, in which a passive soft finger navigates while interacting with plain surfaces and movable objects. Data encompassing proprioception, touch, and vision are recorded within this simulated setting. We explore the fusion of various sensory inputs to construct a coherent perception and present experimental findings in simulation that shed light on the advantages and potential applications of our model within the field of soft robotics, to implement perceptually-aware soft robots.

    \section{Learning Architecture}
        Multi-modal sensory fusion aims to create a comprehensive perception of the environment or the agent's internal state employing a compact state representation, while still retaining relevant and meaningful information. Indeed, different sensory modalities highlight diverse aspects of the same event, and their combination provides a richer representation.

\begin{figure}[t]
          \centering
          \includegraphics[width=\linewidth]{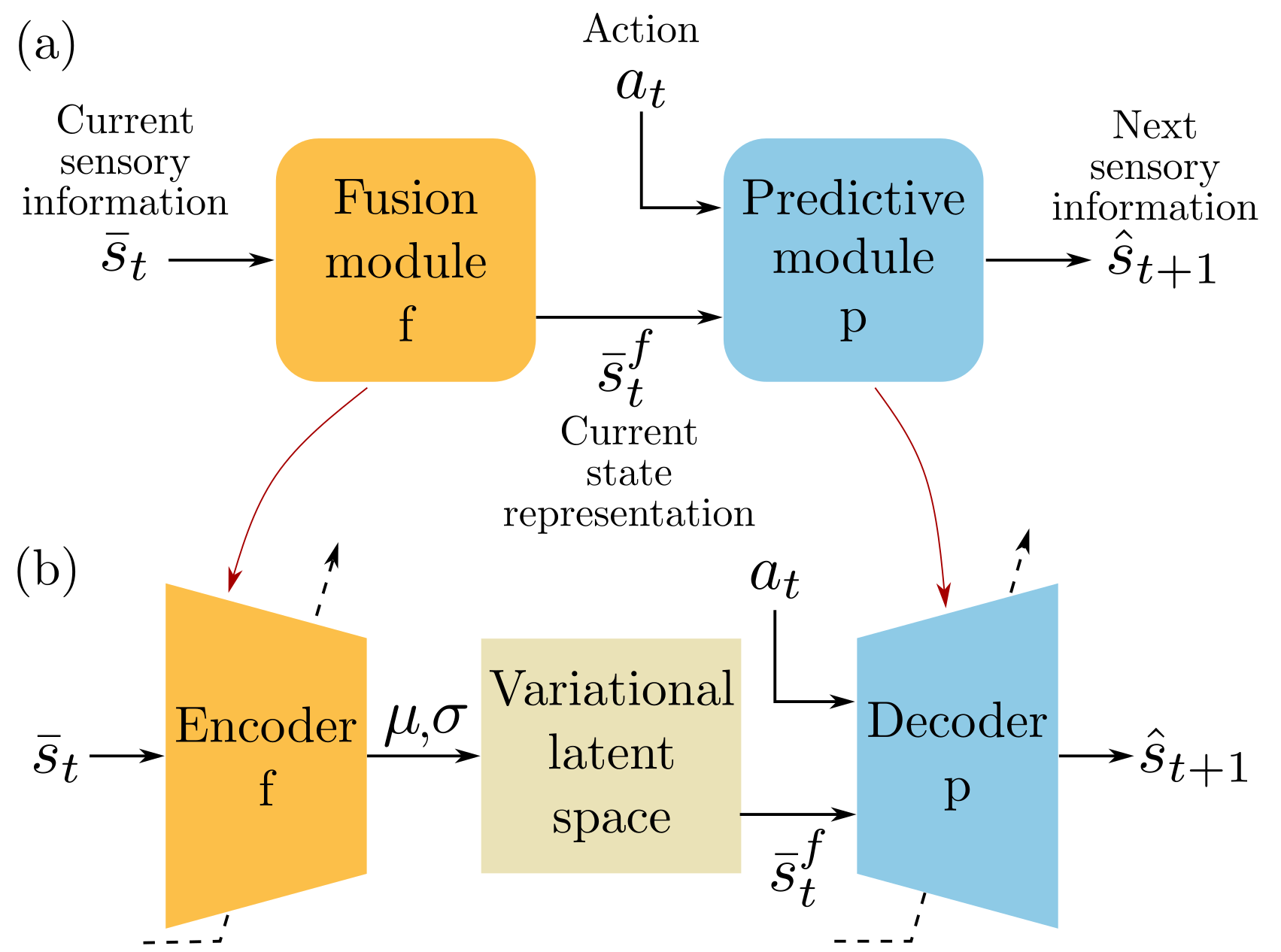}
          \caption{Fusion and prediction learning architecture. (a) Prediction of the next sensory observation starting from a current observation and performed action, after undergoing a fusion and compression stage. (b) Implementation of the model on a Conditional Variational Auto-Encoder.}
          \label{fig:predictive_arch}
        \end{figure}

        The objective of our learning architecture is to facilitate this fusion by capitalizing on the dynamic evolution of the physical body. At any time-step $t$, we can observe a non-minimal variable, denoted as $\bar{s}_t$, from multiple sensory modalities. Given an action $a_t$, these sensory modalities can be combined to predict self or other dependent modalities, $\hat{s}_{t+1}$, in a self-supervised way. Careful consideration of this output sensory information, $\hat{s}_{t+1}$, forces information fusion without manual labelling \cite{lee2019touchvision}. Observations could be composed of sensory inputs coming from different sources of information; in other words, $\bar{s}_t$ and $\hat{s}_t$ are observations of the same event at the same time, but involving different (single, or multiple) sensory modalities. Furthermore, these observations may depend on individual sensory modalities while encompassing distributed information, a key feature that uniquely defines soft robot sensorization. The prediction process can be repeated over consecutive time steps, and the evolution of the state can be summarized as follows:
        \begin{equation}
            \hat{s}_{t+1} = h\left(\bar{s}_t, a_t\right)
            \label{eq:body_dynamics}
        \end{equation}
        Here, the function $h$ represents the dynamics of the body. As it is very unlikely that the raw sensory data, $\bar{s}_t$, constitutes a compact state representation, making long-term predictions is challenging or infeasible. Therefore, there is a need to identify an encoding function $f$ that, based on the output sensory information $\bar{s}_t$, can generate a compact task-independent multi-modal state representation $\bar{s}_t^f$ as expressed in the following equation:
        \begin{equation}
            \bar{s}_t^f = f\left(\bar{s}_t\right)
            \label{eq:compact_representation}
        \end{equation}
        The combination of Eq. \ref{eq:body_dynamics} and Eq. \ref{eq:compact_representation} results in the following identity and it is represented in Figure \ref{fig:predictive_arch}(a).
        \begin{equation}
            \hat{s}_{t+1} = h\left(\bar{s}_t^f, a_t\right) = h\left(f\left(\bar{s}_t\right), a_t\right)
            \label{eq:encoding_decoding}
        \end{equation}

        From a learning perspective, the framework can be implemented using a generative model. Among the various solutions available in the literature, the \ac{VAE} \cite{kingma2013autoencoder} is capable of learning a probabilistic mapping from data to a lower-dimensional space. This aligns with the goals of multi-modal sensory fusion, as it involves encoding a substantial amount of multi-modal information into a smaller representation in a latent space while maintaining completeness and ensuring high compression. However, in our case, the prediction of the next state representation is conditioned by the action undertaken by the system, which helps the decoding stage to better generalize over learning. To summarize, the \ac{CVAE} \cite{sohn2015cvae} in Figure \ref{fig:predictive_arch}(b) implement both sensory fusion and sensory prediction.

        The \ac{CVAE} comprises an encoding and a decoding stage. The fusion encoder $f$ takes the input observation and maps it into a latent probabilistic distribution, assumed to be normal, with mean $\mu$ and standard deviation $\sigma$. A point in the latent space is then sampled stochastically, introducing an element of randomness. The predictive decoder $p$ gets as input the sampled point and the conditional input and attempts to generate data that closely resembles the output data. 

        From a data usage point of view, the network can receive inputs and produce outputs that are either entirely distinct in terms of the modality of information they entail or partially derived from transformed input data. 
        
        The training of such a network involves maximizing a probabilistic objective function, which consists of two parts: reconstruction loss, to minimize the discrepancy between the desired output and generated data, and a regularization term, that encourages the latent space distribution to be close to a standard normal distribution. During training, the model learns the optimal parameters for both the encoder and decoder networks. The encoder learns to produce the mean and standard deviation for the latent space distribution, while the decoder learns to generate data that minimizes the reconstruction loss.

        \begin{figure}[tb]
          \centering
          \includegraphics[width=\linewidth]{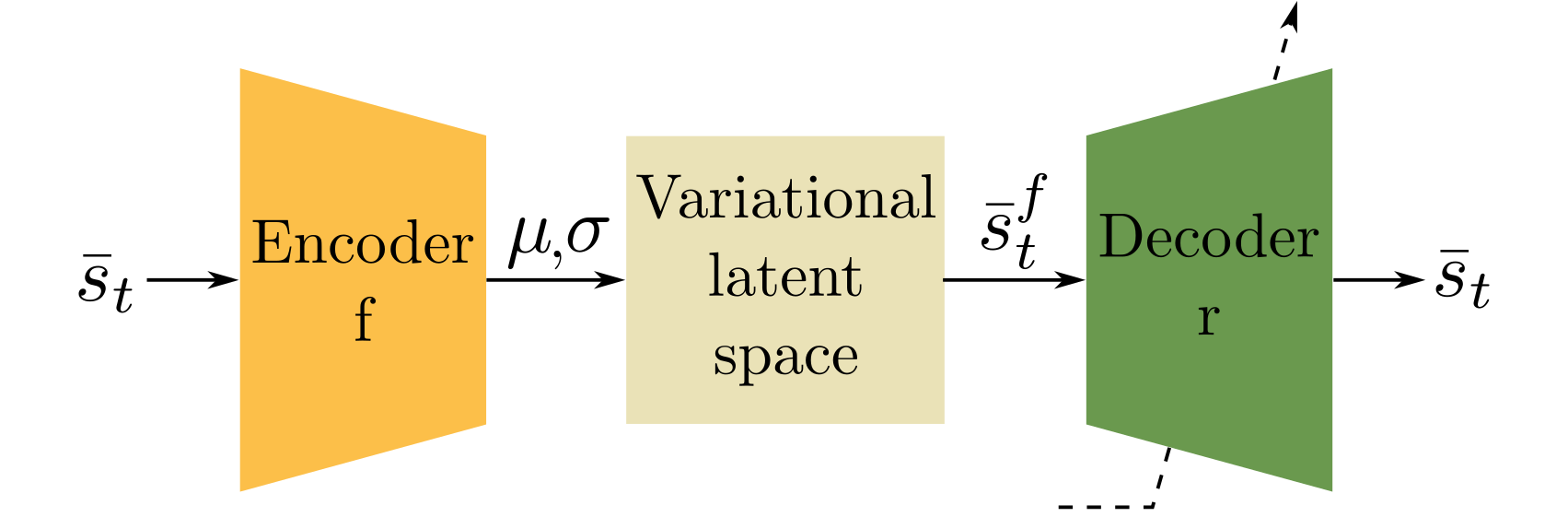}
          \caption{Information reconstruction learning architecture.}
          \label{fig:information_loss_arch}
        \end{figure}
        
        The \ac{CVAE} illustrated in Figure \ref{fig:predictive_arch}(b) can then be utilized to forecast the next sensory observation. This predictive capability serves as a metric for evaluating the network's performance in anticipation. Furthermore, in Figure \ref{fig:information_loss_arch}, the reconstruction decoder $r$ will be utilized to evaluate the informativeness and completeness of the trained state representation in reconstructing the original data after compression. The network's performance will predominantly vary based on the size of the variational latent space. The fusion encoder will not undergo training, but it will be used to map the input observation into the latent space. The reconstruction decoder will learn how to build again original data from the latent representation through self-supervision.
    
    \section{Soft Body Simulation}
        The simulation is realized using the SoMo library \cite{graule2021somo}, which is built upon the PyBullet \cite{coumans2020pybullet} physics simulator. This environment facilitates the emulation of soft-bodied robots, approximated by rigid links interconnected through spring-loaded joints, while effectively handling interactions between the robot's body and the surrounding environment.

        \subsection{Simulation Environment}
\begin{figure}[tb]
              \centering
              \includegraphics[width=\linewidth]{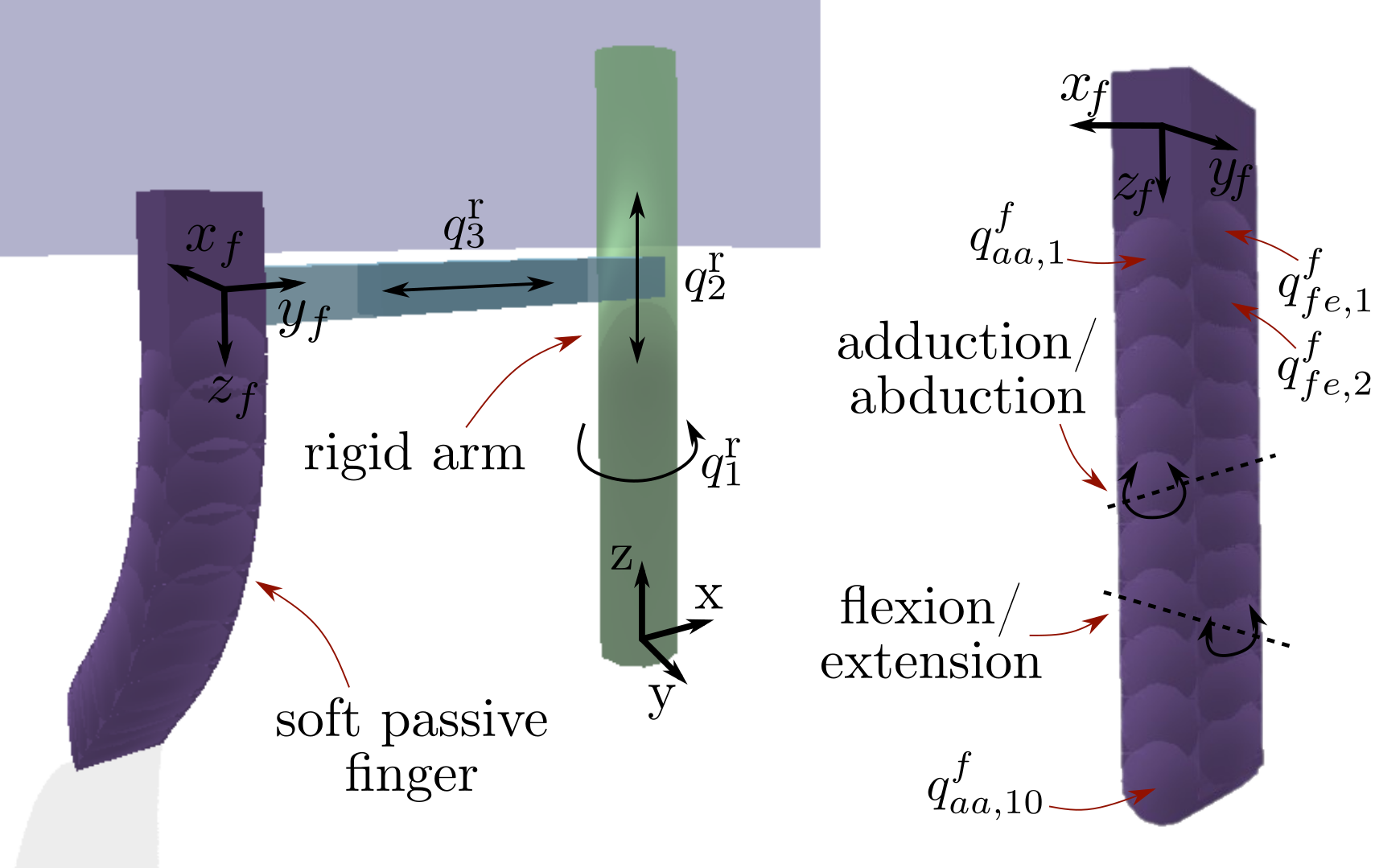}
              \caption{Robotic platform simulation setup. The passive finger is mounted at the distal end of a cylindrical rigid robot and interacts with the ground, or eventually movable objects. The finger presents 20 DoFs and it makes flexion/extension and adduction/abduction movements.}
              \label{fig:sim_env}
            \end{figure}

            The soft passive finger, shown in Figure \ref{fig:sim_env}, comprises a total of 20 links and joints. Each joint possesses an identical spring constant, and all the links share the same mass. The joints are arranged with alternating axes from the base to the tip, enabling the finger to flex in two primary directions: forward and backward, as well as laterally. These movements closely mimic the flexion/extension and adduction/abduction motions observed in the biological finger.
            
            The finger is mounted at the distal end of a cylindrical rigid robot, featuring a rotary joint $q_1$ at its base and two prismatic joints $q_2$, $q_3$ to connect its links. This design provides the robot with a cylindrical-shaped workspace, achieved through a rotating shaft and an extendable arm that moves vertically and in a sliding motion.
            
            Extra objects are introduced into the simulation, as in Figure \ref{fig:task}. Multiple box-shaped objects can be randomly positioned in proximity to the robot, enabling the finger to make additional contact with its surroundings.

        \subsection{Data collection}
            Two distinct types of simulations are conducted to create the training datasets, differentiating between \textit{empty} and \textit{cluttered} scenarios. The distinction lies in the presence of additional objects within the simulation environment. The process initiates with a predetermined initial configuration of the rigid robot. Subsequently, random actuations are generated over a defined number of time steps. This action is intended to cover the entire workspace of the robot arm, thereby enabling flexible finger movements and interactions. The simulation operates at a frequency of 1 KHz to mitigate potential numerical instabilities.

            At each time-step $t$, for both simulation scenarios, a variety of data modalities is recorded, encompassing proprioceptive, contact, and visual information. These data streams provide comprehensive insights into the behaviour and interaction of the system. The sampling time for data acquisition is set to 10 Hz; it ensures a quasi-static motion of the finger.
            
            In the case of proprioceptive data, the angles of the finger joints $q_t^f = \{q_{fe,t}^f, q_{aa,t}^f\} \in \Re^{20 \times 1}$ are logged to capture the finger configuration. This information is crucial for representing the finger's shape. The joint angles of the rigid arm $q_t^r \in \Re^{3 \times 1}$ are also recorded.
            
            Throughout interactions within the simulation, the finger can undergo deformation at either a local or global level. At the local level, PyBullet monitors the interactions occurring between the finger and external objects, and the ground. Specifically, it tracks the normal forces $f_t \in \Re^{20 \times 1}$ acting on each link of the finger. Alternatively, global deformations are captured through visual information. Such visual data $v_t$, with shape $64 \times 64 \times 3$, is obtained by recording the simulation with a virtual camera, assessing the system's overall deformations and interactions from a broader perspective. 
    
    \section{Results} 
        The training datasets have been generated within the simulation environment. It allows for testing our learning architectures on sensing information devoid of issues typical in actual physical systems, such as noise and drift-prone sensing technologies, or the segregation of sensory readout from external force and deformation provided by the actuation system \cite{hu2023eskin}. Future works will explore employing pre-existing datasets or simplified physical robots to benchmark the model on real-world scenarios.
        
        In our experiments, we focus on fine-tuning the hyper-parameters of the learning algorithm through trial and error. A particular focus is placed on the sensory modalities of the input and output data, as well as the dimension of the latent space. Such evaluation takes into account both prediction and reconstruction capabilities.
    
        \subsection{Datasets and training algorithms}
            Two distinct simulations are conducted to assess the performance of the robot in both \textit{empty} and \textit{cluttered} environments. In each simulation, we generated a set of 4k joint actuations for position control, starting from the rigid robot's rest configuration. Rigid robot joints have been randomly moved in these ranges: $q_1^r \in [3\pi/4,5\pi/4]$ [rad], $q_2^r \in [-1,0]$ [m], $q_3^r \in [0,1.5]$ [m]. To ensure smooth transitions between consecutive actuation commands and reduce abrupt changes, we employed a smooth step. We recorded a total of 40k samples for each scenario; each sample contains information about the executed action, the finger's joints configuration, the forces applied to the finger at the link level, and a camera recording capturing the scene.

\begin{figure}[tb]
              \centering
              \includegraphics[width=\linewidth]{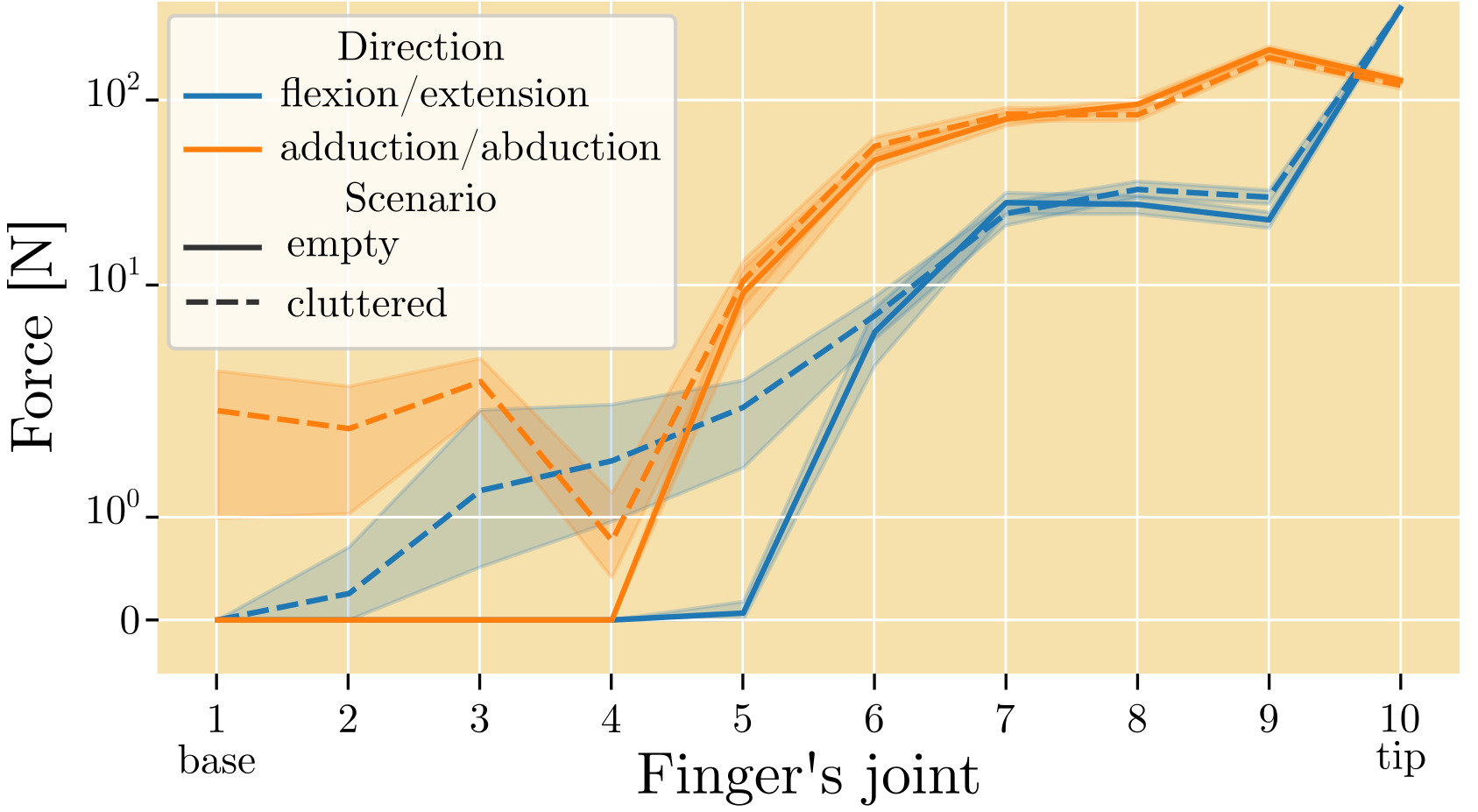}
              \caption{Contact forces in simulation.}
              \label{fig:forces_dataset}
            \end{figure}

            We conducted a detailed analysis of forces, paying special attention to their distribution over the finger's length. This analysis is illustrated in Figure \ref{fig:forces_dataset}. In all scenarios, contact forces primarily affect the distal part of the finger. The presence of objects in the cluttered scenario results in larger forces, applied also to joints proximal to the base.

            Prediction and reconstruction learning architectures share a common implementation, differing primarily in the inclusion of a conditional input during the decoding stage and variations in the input and output cardinality. In the context of single-to-multi-modality prediction, both the encoder and decoder employ two fully connected layers with ReLU activation functions. In the case of multi-to-multi prediction, the encoder begins with a convolutional layer featuring a 4-unit kernel and a 2-unit stride, followed by two fully connected layers, all using ReLU activation functions. A symmetrical structure applies to the decoding stage. The size of the latent layer is selected from a set of options: 2, 4, and 16 variables for the first architecture, and 16, 64, and 128 variables for the second architecture, catering to varying degrees of data compression, from very high to medium and low compression.
            
            Hyper-parameters for the training of the architectures are set to a learning rate of 1e-3, a batch size of $1024$, and a maximum number of epochs set to $200$. Notably, the loss function for the fusion and prediction architecture depicted in Figure \ref{fig:predictive_arch} is the sum of \ac{MSE} and \ac{KLdiv}, which serves to evaluate both reconstruction accuracy and regularization. In contrast, the reconstruction learning architecture solely employs \ac{MSE} to facilitate input and output matching.

            In the context of multi-modal sensory inputs, we combined both vision and proprioception data to serve as feed to the learning architecture. The camera's perspective significantly influences the information conveyed to the perception model. Choosing between egocentric and exocentric viewpoints, we have selected the latter. This entails utilizing a single external camera to oversee the entirety of the rigid arm's workspace and observe interactions between the soft fingers and either the ground or objects. Future works will explore incorporating visual data from the robot's internal perspective, akin to biological systems. This approach introduces challenges such as compensating for self-motion and coordinating visual and proprioceptive information. Additionally, it involves working with incomplete visual data and active vision.
            
        \subsection{Single-to-Multi modality Prediction}
            The predictive architecture is first employed to build a model to map the finger configuration at time $t$ to the configuration and force at time $t+1$. More formally, we will employ an input $\bar{s}_t = q_t^f$ and an output $\hat{s}_{t+1} = \{q_{t+1}^f, f_{t+1}\}$. It is worth noting that the network has to demonstrate predictive capabilities with cross-modal inference, to map the information from the proprioceptive to the contact domain.

            For proprioception forecasting, we employ the \ac{SMAPE} metrics for its evaluation. It provides an estimation of the percentage difference between predicted and actual values in a symmetric way, meaning it doesn't favour overestimations or underestimations. A lower SMAPE indicates a more accurate forecasting by the model. Considering $n$ samples, and for each sample a prediction $ \Tilde{\hat{s}}_t$ of the observation $\hat{s}_t$:
            \begin{equation*}
                \textrm{SMAPE} = \dfrac{100}{n} \sum_{t=1}^n \dfrac{|\hat{s}_t - \Tilde{\hat{s}}_t|}{|\hat{s}_t| + | \Tilde{\hat{s}}_t|}
            \end{equation*}
 
            \begin{figure*}[tb]
              \centering
              \includegraphics[width=\textwidth]{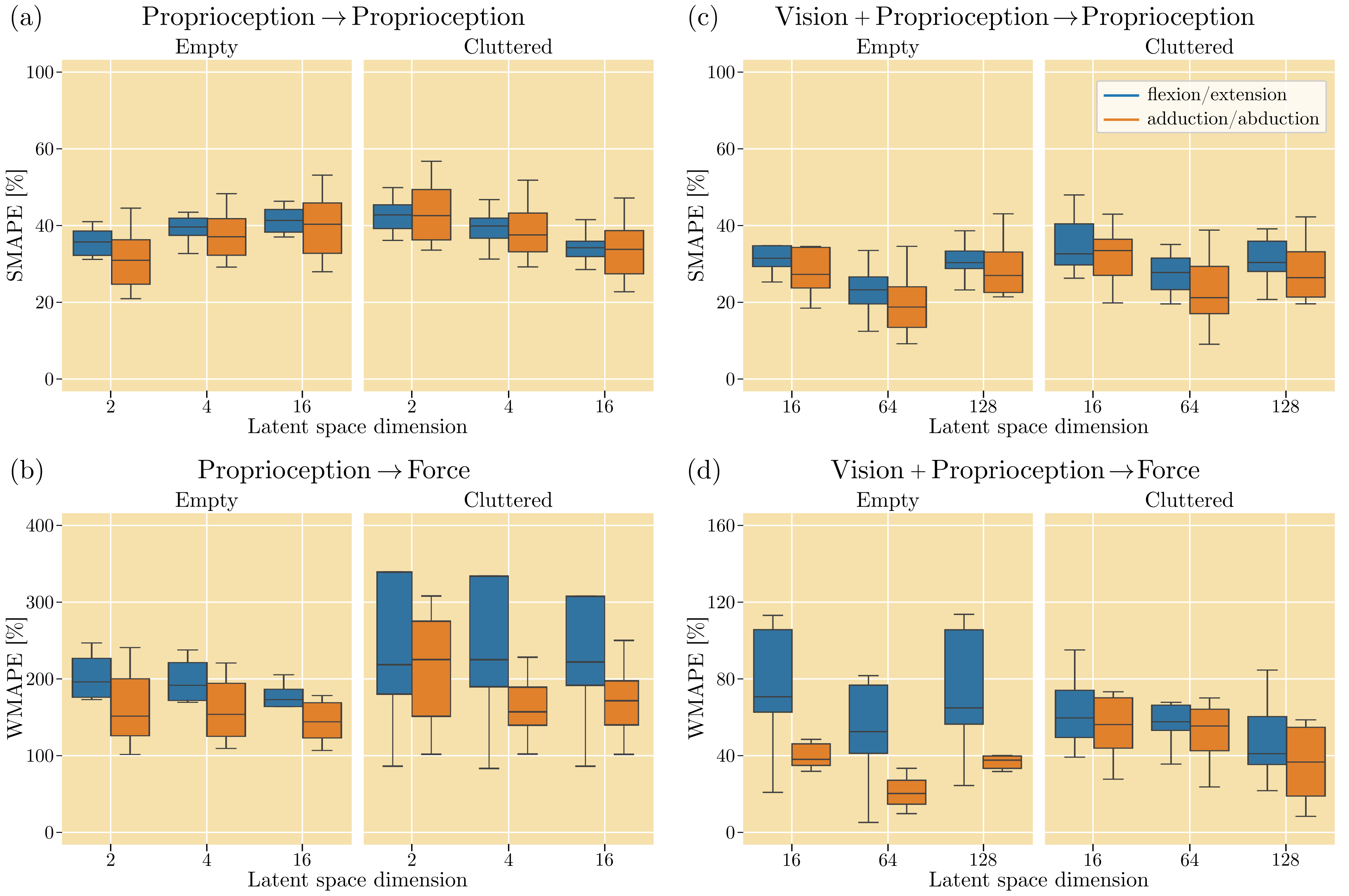}
              \caption{Performance of the Fusion and Prediction learning architecture in both empty and cluttered scenarios, while changing the latent space dimension. (a) Proprioception forecasting over itself. (b) Force forecasting from proprioception. (c) Proprioception forecasting over itself and vision. (d) Force forecasting from the fusion of proprioception and vision.}
              \label{fig:prediction_results}
            \end{figure*}
                      
            Proprioception prediction over a variable latent space dimension is presented in Figure \ref{fig:prediction_results}(a). In both scenarios, lateral movement is more accurately predicted than forward/backward motion. In the empty case, a smaller latent space provides sufficient information for accurate sensory predictions. The limitation in achieving higher dimensions is attributed to the challenge of prediction to converge when dealing with more sparse information. Conversely, in the cluttered scenario, effective forecasting of both degrees of freedom requires less compressed input sensing due to the presence of random obstacles, demanding a higher-dimensional latent space representation.

            Forces values tend to be large, as shown in Figure \ref{fig:forces_dataset}, in which case SMAPE is a poor performance index. Hence, \ac{WMAPE} is introduced, which follows the same considerations as SMAPE.
            \begin{equation*}
                \textrm{WMAPE} = \dfrac{100}{\sum_{t=1}^n |\hat{s}_t|} \sum_{t=1}^n |\hat{s}_t - \Tilde{\hat{s}}_t|
            \end{equation*}
            
            The findings from Figure \ref{fig:prediction_results}(b) clearly indicate that proprioception alone does not provide sufficient information for accurate force prediction. In both scenarios, errors remain consistently high across all latent space dimensions, demonstrating the inability to create a model capable of mapping proprioception to future force, even when conditioned on the specific action employed.

            From a learning perspective, the latent space dimension has a significant impact on the necessary training resources. Generally, a larger latent size corresponds to a greater demand for spatial resources. In an empty scenario, the predictive architecture converges in half the number of epochs when using a medium compact representation compared to other options. Conversely, in a cluttered scenario, a higher dimension results in faster training convergence.

        \subsection{Multi-sensory Fusion and Multi-sensory Prediction}

            \begin{figure}[b!]
              \centering
              \includegraphics[width=\linewidth]{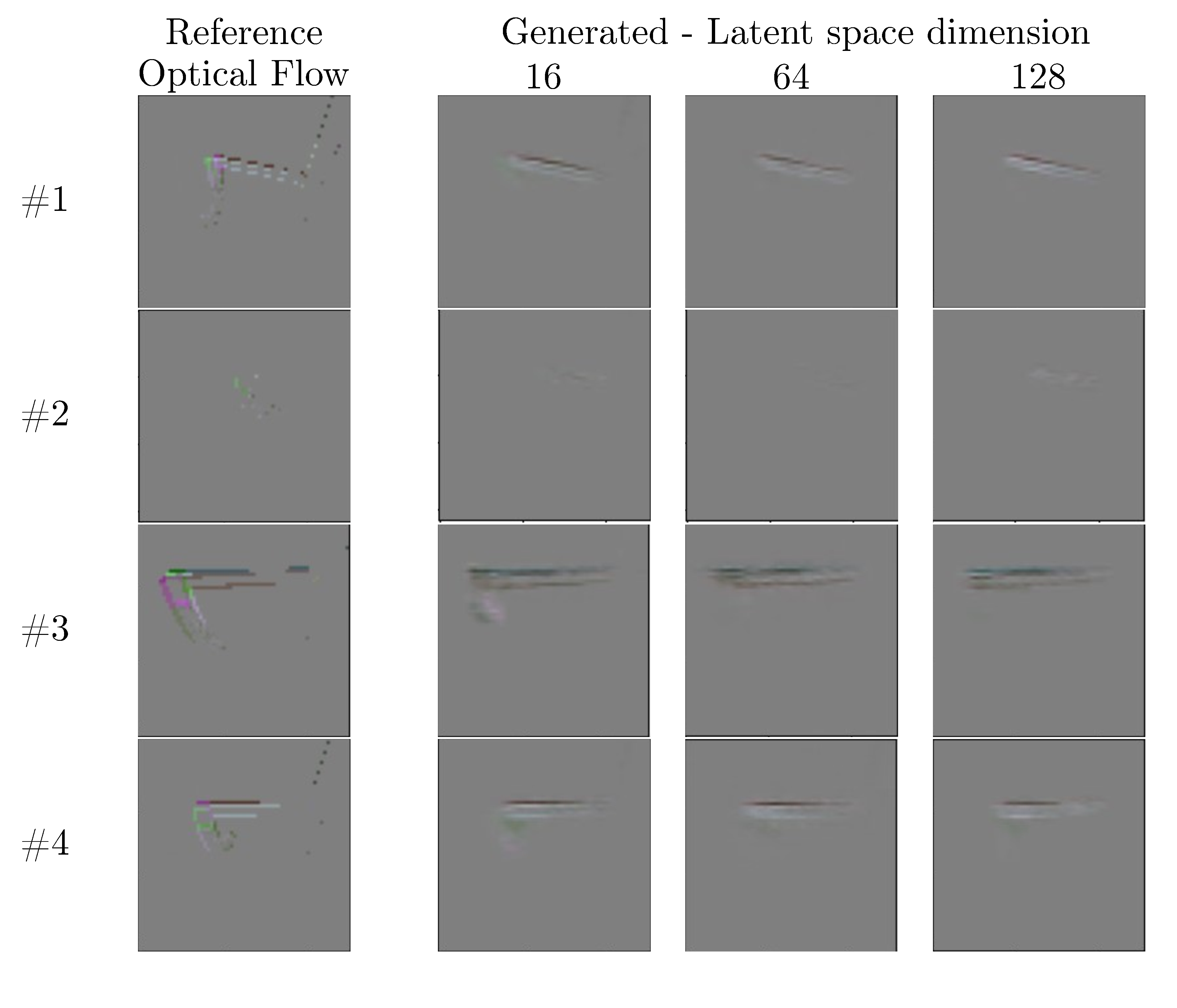}
              \caption{Visual prediction in the cluttered scenario while changing the latent space dimension. Reference optical flow prediction is reported for comparison.}
              \label{fig:propriovision_to_vision}
            \end{figure}
            
            Proprioception can be paired with vision sensing to implement input multi-modality, moving towards our main objective. In this case, from an input $\bar{s}_t = \{q_t^f,v_t\}$ we aim to predict the optical flow $dv_{t+1} = v_{t+1}-v_{t}$ as well as next proprioception and force: $\hat{s}_{t+1} = \{q_{t+1}^f, f_{t+1}, dv_{t+1}\}$.

            When it comes to proprioception, multi-modal prediction surpasses single-modality prediction in both flexion/extension and adduction/abduction. As illustrated in Figure \ref{fig:prediction_results}(c), it becomes evident that a moderate latent layer size yields the best performance in both scenarios. The extremely compressed data fails to retain the essential information, while the least compression does not offer a sufficiently compact state representation for effective prediction. In all scenarios, the greatest errors have been observed at the fingertip level, primarily due to its lack of constraints and the inherently less predictable nature of its behaviour. In essence, this suggests that the integration of vision and proprioception is important for the perception model, even enhancing predictive proprioception itself.
            
            Furthermore, the utilization of multi-modal input enables a substantially higher level of accuracy in predicting the force compared to the single-modality case. The fusion of vision and proprioception offers a wealth of information necessary for this mapping, which can later be leveraged for contact-rich tasks. As depicted in Figure \ref{fig:prediction_results}(d), in the empty scenario, a moderately compact representation offers the most accurate predictions, aligning with the observations made in the proprioception case. Conversely, a larger latent space is necessary in the cluttered scenario due to the challenge involved in contact forecasting with movable objects.     

            Similar considerations apply to training resources as they do to single-modality models: a larger latent space results in greater spatial resource demands, but simultaneously reducing the time required for the predictive model to converge.

            Regarding visual forecasting, Figure \ref{fig:propriovision_to_vision} displays the predictions made across various latent space dimensions, specifically in the cluttered scenario. The most accurately predicted section across all architectures pertains to the motion of the rigid arm, as it is prominently evident in the generated images. Additionally, in trial \#2, it can be observed that the network respects the absence or minimal motion by generating a plain optical flow prediction. However, the most challenging part to predict involves the motion of the soft finger, which is either partially reconstructed (with low-dimensional latent space) or absent. One potential limitation contributing to the network's inability to make accurate predictions could be attributed to the limited depth of the decoder stage, which restricts its ability to retain significant prior scene knowledge, or the sampling frequency of the data.

        \subsection{Information reconstruction}
            The fusion and predictive architecture implements a methodology to gather a condensed state representation that must be enough informative to predict a future observation given the performed action. The amount of compression, given by the dimension of the bottleneck layer, should be large enough to generate a minimal state representation, yet low enough to entail the complete information. While the former bound has been discussed with the employment of the predictive architecture, the latter is assessed through a process of input reconstruction. Using the architecture depicted in Figure \ref{fig:information_loss_arch}, we perform a reconstruction of $\hat{s}_t$ while varying the dimension of the latent layer. 
            
\begin{figure*}[tb]
              \centering
              \includegraphics[width=\linewidth]{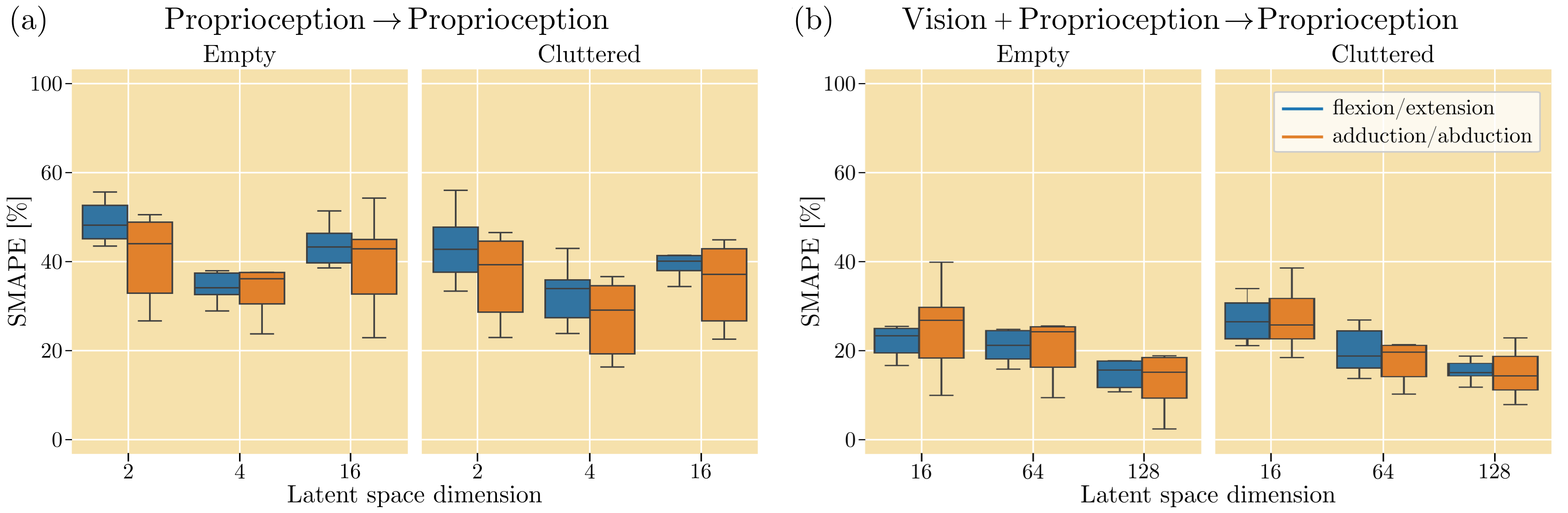}
              \caption{Performance of the Information Reconstruction architecture in both empty and cluttered scenarios, while changing the latent space dimension. (a) Proprioception reconstruction after compression with single-to-multi modality prediction. (b) Proprioception reconstruction after compression with multi-modality prediction.}
              \label{fig:reconstruction_performance}
            \end{figure*}

            Regarding the single modality input, Figure \ref{fig:reconstruction_performance}(a) illustrates that a medium-sized bottleneck layer offers the best level of compression to facilitate further reconstruction. While it is possible to achieve higher levels of compression, they often result in diminished performance, particularly in reconstructing the forward/backward finger movement. Conversely, lower levels of compression do not offer significant advantages over the higher ones, thereby introducing variability in error without substantial performance improvement.

            \begin{figure*}[tb!]
              \centering
              \includegraphics[width=\linewidth]{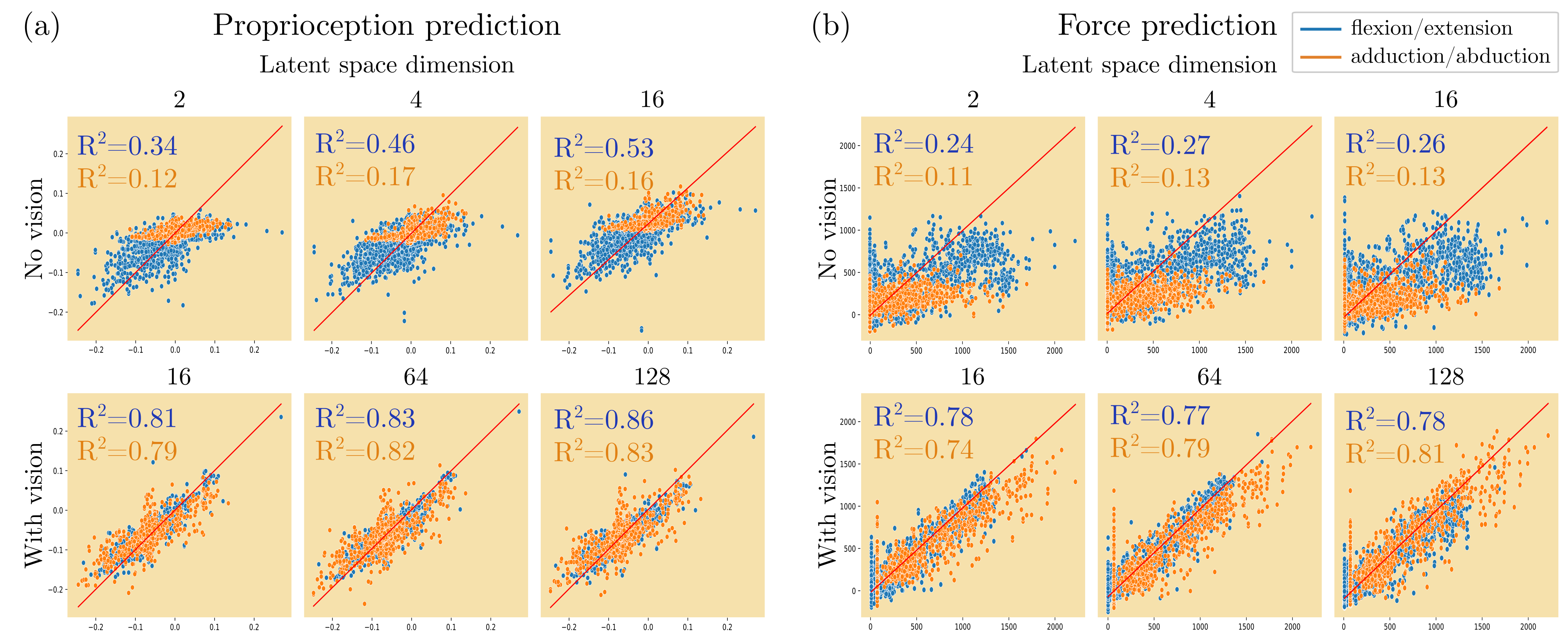}
              \caption{Prediction of (a) proprioception and (b) force at the fingertip compared with reference values, while employing single- or multi-modal inputs and different dimensions of the latent space. The coefficient of determination R2 for both the DoFs has been reported.}
              \label{fig:regression}
            \end{figure*}

            Turning our attention to the multi-modal input, the task of input reconstruction becomes notably more challenging. This is because the state representation involves not only compression but also fusion of information from multiple modalities. As depicted in Figure \ref{fig:reconstruction_performance}(b), the reconstruction performance is, in fact, superior to the single-modality case. This outcome underscores the significant advantage of integrating both proprioception and vision, as it yields a state representation that not only enhances predictions of future observations but also faithfully reconstructs the input. Remarkably, in both scenarios, the most effective reconstruction performance is achieved with less compressed data, even though higher compression levels do not substantially increase the reconstruction error.

        \subsection{Summary}
            Results have shown how to implement the choice for a proper data compression that enables fusion, prediction and reconstruction capabilities through a generative model. These findings are particularly relevant in scenarios involving interactions with movable objects, which further challenge prediction accuracy.

            Regarding proprioception, different insights have emerged. In self-prediction scenarios, a higher compression level is acceptable when dealing with empty environments, but it comes at the cost of poorer reconstruction, necessitating a more moderate compression for efficiency. The presence of objects, on the other hand, introduces the complexity of contact-induced shape changes, demanding a less compressed representation to account for various potential variations. Fusion with vision significantly influences the performance, requiring a less compressed data representation for accurate predictions of future observations and faithful reconstruction across all scenarios. The challenge of selecting the appropriate latent dimension becomes more apparent when we consider the ability to predict the proprioceptive value of the fingertip, which is particularly difficult due to the absence of constraints and its early interaction with the environment. In Figure \ref{fig:regression}(a), the regression performance of the predictive architecture is illustrated when using or omitting visual information in a cluttered scenario. Sensory fusion enhances the regression of proprioceptive prediction as reported by the coefficient of determination R2 for both DoFs, reinforcing the validity of the considerations above.

            One of the most noteworthy results pertains to touch prediction. Solely relying on proprioception proves insufficient for accurately estimating forces, and forces experienced on the lateral side of the finger are generally less accurately predicted than those in the other direction. However, when combined with vision, it offers enough information for robust force estimation in both empty and cluttered scenarios, albeit with a medium to low compression requirement. Figure \ref{fig:regression}(b) visually represents the concept we just discussed through regression analysis. The deficiency in prediction for single-modality input is starkly evident in the top row, but when vision is incorporated, the regression becomes substantially more effective, as it can be observed by R2 improvement, and exhibits improved precision as the latent space dimension increases.
            
    \section{Conclusion}
        We present a perception model designed for soft-bodied robots to construct a concise representation of their sensory experiences. This achievement hinges on the development of a predictive model that operates across various sensory modalities, offering insights into the upper bound of information compression. Simultaneously, we employ an information reconstruction model to establish a lower bound. These boundaries inform the selection of a compact yet comprehensive internal model, enabling the implementation of more advanced control strategies.

        Our model demonstrates its adaptability by effectively managing diverse sensory modalities, including their fusion and prediction. Future work will delve into exploring the impact of early and late fusion, as well as early and late prediction, on the model's performance. We will also investigate the optimal prediction window size to achieve the highest accuracy when transitioning from output to input, ensuring long-term network stability. Additionally, we will consider recurrent models for both input and output to address the inherent hysteresis in soft-bodied robots.
        
        To enhance the scope of the model, we will expand its applicability to include actuated soft robots and evaluate the effects of such inclusions on the perception model. We intend to deploy the network on a real-world soft robot for further in-depth analysis and experimentation. Ultimately, we aim to use this compact and task-independent state representation for developing task-specific control policies using reinforcement learning methods. 
        
    \printbibliography
\end{document}